\documentclass[letterpaper]{article} 
\usepackage{main}  
\usepackage{times}  
\usepackage{helvet}  
\usepackage{courier}  
\usepackage[hyphens]{url}  
\usepackage{graphicx} 
\usepackage{natbib}  
\usepackage{caption} 
\usepackage{algorithm}
\usepackage{algorithmic}
\usepackage{comment}

\usepackage{amsmath}
\usepackage{amssymb}

\usepackage{newfloat}
\usepackage{listings}
\DeclareCaptionStyle{ruled}{labelfont=normalfont,labelsep=colon,strut=off} 
\floatstyle{ruled}
\newfloat{listing}{tb}{lst}{}
\floatname{listing}{Listing}
\nocopyright 

\title{Comparative Analysis of Multi-Agent Reinforcement Learning Policies for Crop Planning Decision Support}
\author{
    Anubha Mahajan\textsuperscript{\rm 1}\equalcontrib,
    Shreya Hegde\textsuperscript{\rm 2}\equalcontrib,
    Ethan Shay\textsuperscript{\rm 3},
    Daniel Wu\textsuperscript{\rm 4},
    Aviva Prins\textsuperscript{\rm 3}
}
\affiliations{
    \textsuperscript{\rm 1}Georgia Institute of Technology, Atlanta, GA, USA\\
    \textsuperscript{\rm 2}Amherst College, Amherst, MA, USA\\
    \textsuperscript{\rm 3}University of Maryland, College Park, MD, USA\\
    \textsuperscript{\rm 4}Montgomery Blair High School, Silver Spring, MD, USA\\
    \textsuperscript{\rm 1}amahajan68@gatech.edu,
    \textsuperscript{\rm 2}shegde26@amherst.edu,
}

\begin{document}

\maketitle

\begin{abstract}
In India, the majority of farmers are classified as small or marginal, making their livelihoods particularly vulnerable to economic losses due to market saturation and climate risks. Effective crop planning can significantly impact their expected income, yet existing decision support systems (DSS) often provide generic recommendations that fail to account for real-time market dynamics and the interactions among multiple farmers. In this paper, we evaluate the viability of three multi-agent reinforcement learning (MARL) approaches for optimizing total farmer income and promoting fairness in crop planning: Independent Q-Learning (IQL), where each farmer acts independently without coordination, Agent-by-Agent (ABA), which sequentially optimizes each farmer's policy in relation to the others, and the Multi-agent Rollout Policy, which jointly optimizes all farmers' actions for global reward maximization. Our results demonstrate that while IQL offers computational efficiency with linear runtime, it struggles with coordination among agents, leading to lower total rewards and an unequal distribution of income. Conversely, the Multi-agent Rollout policy achieves the highest total rewards and promotes equitable income distribution among farmers but requires significantly more computational resources, making it less practical for large numbers of agents. ABA strikes a balance between runtime efficiency and reward optimization, offering reasonable total rewards with acceptable fairness and scalability. These findings highlight the importance of selecting appropriate MARL approaches in DSS to provide personalized and equitable crop planning recommendations, advancing the development of more adaptive and farmer-centric agricultural decision-making systems.
\end{abstract}

%

\section{Introduction}
\label{sec:intro}

India’s agricultural sector faces major challenges. Over half of agricultural households are in debt, often due to crop failures \citep{statista_agri_debt}. Effective crop planning is key to improving farmers' incomes, but existing systems do not adapt well to real-time market supply and demand. This often leads to income loss, especially when recommendations are homogenous, causing market saturation and lower profits. For small-scale farmers, particularly in countries like India, crop planning that adjusts dynamically can significantly boost their earnings \citep{Fabregas2019}.

Crop planning is a crucial challenge for small-scale farmers, especially in developing nations like India, where they face significant risks from climate variability, market fluctuations, and limited resources \citep{matthan2023beyond}. Decisions on what to plant, when to harvest, and how to manage resources are critical to farmers’ livelihoods, yet traditional decision-support systems (DSS) often fall short of addressing the complexities they face. These systems do not account for the interconnected nature of farming communities, where one farmer's choices can impact others, especially in shared markets where oversupply can drive down prices. As a result, farmers are left vulnerable to economic instability and are often unable to achieve equitable outcomes \citep{esteso2022optimization}. 

While reinforcement learning (RL) has demonstrated success in agricultural applications such as fertilization management policies \citep{Overweg2021}, its potential for crop planning remains underexplored, particularly in multi-agent settings. Traditional RL systems primarily focus on optimizing outcomes for a single agent, overlooking the critical role of coordination among multiple farmers. Multi-agent reinforcement learning (MARL) offers a promising framework for addressing this gap, enabling farmers to optimize their actions collectively while managing shared resources and mitigating market risks. Recent advancements in MARL highlight its application in tasks such as spatiotemporal sensing \citep{Tamba2022} and task offloading and network selection in heterogeneous agricultural networks \citep{Zhu2023}. Additionally, MARL has been used to coordinate UAVs for field coverage and crop monitoring \citep{Marwah2023}. Despite these innovations, existing MARL applications often focus on objectives like spacial coverage or resource allocation, overlooking market saturation and dynamic interdependencies in crop planning. This highlights the need for MARL systems that enhance efficiency, include agent interactions, address market dynamics, and promote economic stability in shared decision-making.

In this paper, we explore how three different MARL algorithms manage joint rewards and fairness in greenhouse crop planning in Telangana, India: (a) Independent Q-Learning \textsc{(IQL)}, (b) Agent-by-Agent \textsc{(ABA)}, and (c) Multi-agent \textsc{Rollout} Policy. Each algorithm has a unique approach: \textsc{IQL} allows each agent to learn independently, optimizing for individual rewards without directly considering the actions of others, which simplifies computation. ABA takes a sequential approach, optimizing the policy of one agent while holding the others constant, improving coordination across agents while maintaining manageable computational complexity. Finally, \textsc{Rollout} takes a more comprehensive approach by considering the joint actions of all agents at each step, aiming to maximize total collective rewards through simultaneous optimization. 

We conducted experiments to analyze the performance of these algorithms, focusing on key metrics such as runtime, total joint reward, and aggregated welfare, concepts we will define later. Additionally, we adjusted hyperparameters representing long-term reward importance (discount factor, $\gamma$) and sensitivity to changes in reward distribution (slope coefficient), allowing us to assess stability and fairness across different approaches. The results show that IQL is computationally efficient but struggles to coordinate and maximize joint rewards, while ABA provides a better balance between efficiency and fairness. \textsc{Rollout} achieves the highest joint reward but at the cost of significantly increased runtime. For practitioners seeking an efficient solution with reasonable fairness, ABA offers a balanced approach, whereas those prioritizing maximum joint reward may consider \textsc{Rollout} despite its computational cost. These insights highlight the trade-offs between reward optimization, fairness, and computational complexity in the context of multi-agent crop planning.

\section{Related Works}

Crop planning is a critical issue for small-scale farmers, particularly in developing nations like India where climate variability and market fluctuations can lead to severe economic losses. Various DSS have been developed to assist farmers in selecting optimal planting and harvesting schedules. Traditional approaches, such as those implemented by \citep{Sarker2002, Mohamed2016, Bhatia2020}, utilize rule-based systems or linear programming techniques to provide generalized crop recommendations based on average weather and market conditions. However, these systems fail to account for the heterogeneity of farmers' individual circumstances and the non-stationary nature of both environmental and economic factors.

RL offers a more dynamic approach to crop planning by enabling systems to adapt based on continuous feedback from the environment. For example, \citet{yang2020deep} applied reinforcement learning to optimize irrigation scheduling, achieving improved water-use efficiency in agriculture. Similarly, \citet{chen2021reinforcement} proposed an RL-based system that accounts for weather predictions and soil moisture data to inform farmers' irrigation practices. While effective for single-agent settings, these systems do not account for the complex interactions between multiple farmers, whose simultaneous actions can influence market prices and resource availability.

While MARL holds promise for addressing challenges in collaborative decision-making, its application in crop planning remains limited. Most existing research explores related but distinct areas, reducing its direct applicability. For example, \citet{kamali2024multi} developed a multi-agent simulation to evaluate groundwater management policies, where agents negotiate water usage based on local needs. This study highlights MARL's potential in managing shared resources but does not address crop planning dynamics. Similarly, \citet{Li2018} applied multi-agent systems in agricultural markets to optimize planting schedules based on market conditions. However, their approach primarily focused on price predictions and overlooked the role of agent interactions in mitigating market saturation. These gaps highlight the need for further exploration of MARL frameworks tailored to the complexities of crop planning.

Our work builds on this existing research by investigating three RL approaches within a multi-agent setting to address critical gaps in crop planning strategies, particularly in optimizing joint rewards among farmers. Conventional systems often rely on uniform recommendations; in contrast, our system dynamically adapts to the actions of other agents (farmers) to mitigate market saturation and promote fair distribution of profits. Each policy is evaluated based on its trade-offs in computational efficiency, coordination, and fairness. This analysis uncovers underexplored dynamics in MARL approaches and offers insights into their potential to advance collaborative decision-making in agricultural systems.

\section{Methods}

First, we will discuss how we model the problem. We will then describe the objective which our algorithms seek to optimize. Next, we will describe the three algorithms which we have developed to solve our problem. 

\subsection{Problem Modeling}

As mentioned in Section \ref{sec:intro}, we are investigating greenhouses in Telangana, India. We model the greenhouses as a collection of identical Markov Decision Processes (MDPs). We extend \citet{lu2024online}'s model of individual greenhouse-agents with a new model to capture interactions between agents. We capture the problem over a finite horizon of $T \in \mathbb{N}$ discrete timesteps, where a fixed number of days passes with each timestep.

There are $n$ agents belonging to the agent set $I$. Every agent $i \in I$ owns a greenhouse, whose condition at a given timestep is represented by one of many states. Each state holds information about which crop is planted, and whether it is harvestable. We notate the state of greenhouse $i$ at timestep $t$ as $s_{t, i}$, and the the joint-state of all agents at timestep $t$ as $\vec{s_t} = \langle s_{t,1}, s_{t,2}, ..., s_{t,n} \rangle$. There exists a common state space $S$ such that $s_{t,i} \in S$ for all agents $i$ and timesteps $t$. Each agent $i$ starts at some state $s_{1,i}$, sampled from $\mu_i \in \triangle(S)$.

At every timestep, each agent takes some action. We will notate the action of agent $i$ at timestep $t$ as $a_{t,i}$, and the joint-action of all agents at timestep $t$ as $\vec{a_t} = \langle a_{t,1}, a_{t,2}, ..., a_{t,n} \rangle $. There exists a common state space $A$ such that $a_{t,i} \in S$ for all agents $i$ and timesteps $t$. In particular, there is an action $\text{HARVEST} \in A$.

Because of crop seasonality, the transition function is non-stationary. Specifically, for every timestep $t$, there is a transition function $P_{t}: S \times A \rightarrow S$ common to all agents where state transitions are determined as $s_{t+1,i} = P_t(s_{t,i},a_{t,i})$. Here, the transitions are deterministic. Interested readers can refer to \citet{lu2024online} for detailed descriptions on the states, actions, and transitions for our problem.

We now introduce interactions between agents corresponding to marketplace competition. After every timestep, each agent receives a reward. We notate the reward for agent $i$ at timestep $t$ as $r_{t,i} \in \mathbb{R}$, and the joint-reward of all agents at timestep $t$ as $\vec{r_t} = \langle r_{t,1}, r_{t,2}, ..., r_{t,n} \rangle $. Rewards are non-stationary but not agent-independent. Specifically, for every timestep $t$, there is a reward function $\vec{R}_{t}: S^n \times A^n \rightarrow \mathbb{R}^n$ where $\vec{r_t} = \vec{R}_{t}(\vec{s_t}, \vec{a_t})$. 

As it is specific to market dynamics, our model's reward function is highly structured. First, agents receive rewards only if they make a valid harvest. This means that $\vec{R}_{t}(\vec{s_t}, \vec{a_t})_i > 0$ if and only if $s_{t,i}$ is harvestable and $a_{t,i} = \text{HARVEST}$. 

Secondly, $\vec{R}_t$ is broken down by crop, and rewards depend only on the number of agents who made a valid harvest of each crop. We will define $d_{t,c}$ as the number of agents who made a valid harvest of crop $c$ on timestep $t$. For each timestep $t$ and crop $c$, there is a pseudo-reward function $Y_{t,c}: \mathbb{N} \rightarrow \mathbb{R}$ such that $Y_{t,c}(d_{t,c})$ indicates the market price of crop $c$ at timestep $t$, when $d_{t,c}$ sellers are supplying that crop. $\vec{R}_t$ is the function which results from assigning the appropriate market price to all agents who made a valid harvest. 

$Y_{c,t}(d_{t,c})$ is monotonically decreasing on $d_{t,c}$. This is because as the number of sellers of a crop increases, the price of that crop decreases, as the larger number of sellers will bid down the price through supply-demand mechanics. We modeled $Y$ as a linear function on $d_{t,c}$. We generated the linear fit through a regression of past market data of tuples of the form (crop, kg supplied, price). Specifically, we fitted two parameters $a_{t,c} < 0$, and $b_{t,c} > 0$ and set $Y_{c,t}(d_{t,c}) = a_{t,c}(SLOPE\_COEFFICIENT*d_{t,c}) + b_{t,c}$. The slope coefficient is a hyperparameter which adjusts the market impact of each agent.

\subsection{Objective}

The decision maker's goal is to develop a policy for each agent, which at each timestep takes in a state and returns an action. Formally, the solution will be a collection of mappings $\pi_{i,t}: S \rightarrow A$ for all agents $i$ and timesteps $t$. This indicates that if agent $i$'s greenhouse in state $s$ at time $t$, then we recommend that they take action $\pi_{i,t}(s)$. We will notate the full collection of policies as a joint-policy $\pi$.

Farmers cannot observe what is happening in other greenhouses. Therefore, the objective's type signature is limited to finding independent policies for each agent, meaning that each agent decides actions based only on the state of their own greenhouse, and not on the states of other agents. This choice also has advantages for computational complexity, as there are an exponential order more non-agent-independent policies than agent-independent policies. 

We wish to maximize the total welfare of all agents in our system. In order to define total welfare, we will first define returns for each agent as an expectation of the weighted sum of the rewards received over the horizon, parameterized by some discount factor $\gamma$. The discount factor weights how much present rewards are prioritized over future rewards. The return of agent $i$ is $g_i = \mathbb{E}(\sum_{t=1}^{T} \gamma^{t-1} r_{t,i}$). We will notate the joint-returns as $\vec{g}$, and will sometimes use $g_i(\pi)$ or $\vec{g}(\pi)$, since the returns are fully dependent on $\pi$.

Now, we will define the total welfare of all agents as $U = \prod_{i=1}^{n}(g_i+1$). We choose to optimize the product of returns because it combines a preference for all agents have higher returns, and a preference for returns to be distributed equally. We can also think of this as optimizing $\log(U) = \sum_{i=1}^n(\log(g_i+1))$. This is motivated since many studies have shown that happiness has a logarithmic relationship with income. We will sometimes use $U(\pi)$ notation.

\subsection{Independent Q-Learning}

\textsc{IQL} adapts single-agent reinforcement learning principles to multi-agent systems \cite{tan1993multi}. Each agent independently learns its policy by observing changes in the environment, disregarding the actions of other agents as direct influencing factors. This simplifies the multi-agent problem by treating each agent as if it were operating in a single-agent context, where the focus lies on learning its own Q-function. A Q-function is defined as the expected future rewards for taking a specific action in a given state and following an optimal policy thereafter.

To implement \textsc{IQL}, we construct a Q-table that captures these expected rewards for each possible action in each state. The Q-values in this table are continuously updated as the agent gathers new observations, enabling it to better its decision-making over time. We first choose a learning parameter $\alpha$. Then, for agent 
$i$, the Q-value update rule is defined as:
\begin{align}
    Q_i(s, a_i) &\leftarrow Q_i(s, a_i) + \alpha \big[ r_i + \gamma \max_{a'_i} Q_i(s', a'_i) \notag \\
    &\quad - Q_i(s, a_i) \big].
\end{align}

\begin{algorithm}[tb]
\caption{Independent Q-learning}
\label{alg:IQL}
\textbf{Input}:State space $S$, action space $A$, transition dynamics $P_t$, reward function $\vec{R}_t$, agent set $I$, time horizon $T$, initial state distribution $\mu_i$, discount factor $\gamma$
\textbf{Parameter}: learning rate $\alpha$, exploration probability $\epsilon$\\
\textbf{Output}: Updated Q-values $Q_i(s, a_i)$ for each state-action pair
\begin{algorithmic}[1]
\STATE Initialize $Q_i(s, a_i)$ for all $s \in S$ using LPSolver
\FOR{each episode}
\FOR{each time step $t = 0, 1, 2, \ldots$}
\STATE Observe current state $s_t$
\STATE Choose action $a_{t,i} \in A_i$:
\IF{Random Variable $\sim [0,1] < \epsilon$}
\STATE Select a random action $a_{t,i}$
\ELSE
\STATE Select $a_{t,i} = \arg \max_{a_i} Q_i(s_t, a_i)$
\ENDIF
\STATE Observe actions of other agents $a_{t,j}$ for $j \ne i$
\STATE Observe reward $r_{t,i}$ and next state $s_{t+1}$
\STATE Update $Q$-value:
\STATE $Q_i(s_t, a_{t,i}) \gets Q_i(s_t, a_{t,i}) + \alpha \left( r_{t,i} + \gamma \max_{a_i'} Q_i(s_{t+1}, a_i') - Q_i(s_t, a_{t,i}) \right)$
\ENDFOR
\ENDFOR
\end{algorithmic}
\end{algorithm}

The \textsc{IQLBufferPolicy} algorithm is an extension of the IQL approach that introduces two key improvements: optionally initializing the Q-values by solving a Linear Programming (LP) problem using an offline base policy and sampling multiple trajectories for each agent to enhance the policy's learning process over a longer horizon. Please note that we will use \textsc{IQL} to refer to \textsc{IQLBufferPolicy} from here on. Here’s a brief explanation of the main methods in the IQL code:

\begin{itemize}
    \item The Q-tables for each agent are initialized, and if the warm start flag is set, the Q-values are ``warm-started" using an \textsc{LPsolver} base policy. This offline policy computes the initial Q-values by solving for the agent's expected reward in each state based on the transition matrices and reward function of the MDP model. This step significantly accelerates the learning process by providing an improved starting point for Q-values. Further information about the \textsc{LPsolver} base policy can be found in Appendix C.

    \item In this method, the Q-table is updated over multiple episodes, simulating a trajectory of actions and observing the resulting rewards and state transitions. This allows for more robust learning as the agent samples multiple paths and adjusts its Q-values accordingly.

    \item Choose an action using the epsilon-greedy policy for a specific agent. The epsilon-greedy policy balances exploration and exploitation. With probability $\epsilon$, a random action is chosen to explore new actions. Otherwise, the action with the highest Q-value is chosen to exploit known information.

    \item The Q-value for a state-action pair is updated based on the observed reward and the maximum Q-value of the next state according to the equation shown earlier (following the standard Q-learning update rule). This allows the agent to incrementally improve its policy by considering both immediate and future rewards.

    \item Run the Independent Q-learning algorithm for multiple episodes. In each episode, the agents interact with the environment for a fixed number of timesteps (horizon). The state, actions, rewards, and next states are observed, and the Q-values are updated accordingly.

    \item After the Q-tables are updated, the algorithm computes the best joint policy for the cohort of agents by selecting the action with the highest Q-value for each state and agent. The joint action is computed at each time step, taking into account the current state of all agents.

\end{itemize}

\subsubsection{Time complexity}

IQL simplifies multi-agent reinforcement learning by treating each agent independently. For each episode $ m \in M$, we loop through all timesteps $T$, where each agent selects an action and observes a reward based on the current state. At each timestep, each agent incurs a cost of $ O (|A|+|S|)$ where $A$ represents action space and $S$ the state space. Thus, the overall time complexity, which is linear in the number of agents and timesteps, is $O(MTN|S||A|)$. Unfortunately, although this approach is computationally efficient and easy to implement and understand, there is no guarantee of convergence after $T$ timesteps. This is due to the non-stationary environment created by the independent learning processes of multiple agents.

\subsection{Agent-by-Agent Policy}

The \textsc{ABA} algorithm breaks down the global optimization problem of finding a policy for every agent into multiple local optimization problems of finding the best policy for a single agent, while holding the policies of all other agents constant. This allows us to reduce the multi-agent problem into a single-agent problem, where the $n-1$ agents not actively being optimized are treated as if they were part of the environment, simply affecting the reward function. This agent-by-agent optimization is related to the family of coordinate descent algorithms, in which one parameter is optimized at a time until convergence. In our case, the ``coordinate" directions are the single-agent policies $\pi_i$ for each agent $i$. 

After initialization, the algorithm loops until a policy convergence threshold is reached. Within this outer loop, a single agent $i$ is chosen, and then the algorithm finds the welfare-maximizing policy for agent $i$, according to the currently held policies of all other agents. Formally, we seek to find the single-agent policy $\pi_i ^{*}$ such that $U(\langle \pi_{-i}, \pi_i ^{*} \rangle)$ is maximized. We then update $\pi_i \leftarrow \pi_i^{*}$ and continue to the next iteration, choosing another agent to optimize.

\subsubsection{Dynamic Programming}

The most involved portion of this algorithm comes down to solving for the welfare-maximizing policy for a single-agent, given the policies of all other agents. Our solution takes advantage of the acylicity of timesteps, and uses dynamic programming.

For illustration, suppose that we changed our framing of the problem so that we expand our state space to have a state for every ($s, t$) pair for all $s \in S, 1 \leq t \leq T$. The resulting state transition graph would be acyclic, giving us a topological order with which to process states. If we process states in descending order from $t=T$ to $t=1$, then by the time our algorithm processes a state $s$, we will have already processed all states $s'$ that $s$ can transition to. Further, if we know the value of state $P(s,a)$ for all actions $a$, we can determine the value of each action $a$ as a sum of the instantaneous reward and the value of our new state $P(s,a)$.

Thus, we are yielded the dynamic programming approach. As we are optimizing $\pi$ for some single-agent $i$, we define $DP_{t}(s)$ as the maximal increase to $U$ possible from timestep $t$ onwards, when agent $i$ starts at $s$, and all other agents $j \neq i$ follow $\pi_j$ for every timestep. As stated, we will compute $DP$ values in descending order, so that we have access to values of $DP_{t+1}$ when calculating values of $DP_t$.

Now, we must find transition equations which relate $DP$ values. Because welfare is non-linear on returns, it is impossible to determine the actual effect of taking a certain action without knowing the returns of all agents across the entire horizon, which is dependent on the joint-policy. Because the whole point of our dynamic programming approach is to find the best $\pi_i$, we do not have access to $\pi_i$ when we are computing $DP$ values. Thus, we will have to settle for an approximation of the welfare added from an action, which depends only on the parameters accessible when computing $DP$ values: namely $\pi_{-i}, s, a,$ and $t$. We will do this by linear approximation. 

\subsubsection{Linear Approximation}

Because linearity allows us to isolate the effect of a single reward, the approximation we elect will be a first-order linear approximation of the welfare function evaluated at $\pi$ before the current iteration of single-agent optimization. Specifically, we approximate the change in $U$ under some change $r_{t,j} \leftarrow r_{t,j} + \epsilon$ as $\epsilon * \frac{\partial U}{\partial r_{t,j}} |_{\pi}$. Evaluating the partial derivatives at the previous version of the joint-policy incentives the optimization to improve upon its prior form.

The details of how this is implemented are mathematically involved. Please see the appendix for all of the details on this portion of the algorithm.

\subsubsection{Details}

Before the main portion of the algorithm, each agent's policies must be initialized. This can either be done by choosing an arbitrary action $a \in A$ and set $\pi_{i,t}(s) \leftarrow a$ for all $i, t, s$, or by uniformly randomizing the entire policy table.

Additionally, we need a stopping criterion to exit the algorithm’s outer loop. A simple choice is to run the loop for a fixed number of iterations $e$ where $e$ is a hyperparameter to be optimized empirically. Another option is to set a parameter $\delta$ and terminate when the improvement in welfare from our refined policy is less than $\delta$.

To choose which agent is selected to optimized, two choices stand out:
\begin{enumerate}
    \item Cycle through agents in order $1$ through $n$ until convergence. This approach has the drawback of allowing some systematic limitations; for instance, agent $i+1$ always immediately responds to changes in agent $i$'s policy, whereas the reverse requires iterating through all other agents.
    \item Randomly select one of the $n$ agents at each iteration. Randomness often works well in descent-based optimization algorithms, like stochastic gradient descent. If sufficient training iterations are run, agents will be selected roughly equally with high probability.
\end{enumerate}

When we can't directly access $\vec{R}_t$ for future $t$, we approximate it as $\hat{\vec{R}}_t$ using observations of past $\vec{R}_t$, allowing this algorithm to adapt to real-time market updates. There are many ways to do this which could be explored in future research.

\begin{algorithm}[] 
\caption{Agent-by-Agent Optimization With Dynamic Programming}
\label{alg:AAO_DP}
\textbf{Input}: State space $S$, action space $A$, transition dynamics $P_t$, reward function $\vec{R}_t$, agent set $I$, time horizon $T$, initial state distribution $\mu_i$, discount factor $\gamma$ \\
\textbf{Output}: Policy $\pi$

\begin{algorithmic}[1]
\WHILE{Convergence Criteria Not Met}
    \FOR{$t=1$ to $T$}
        \FOR{$i \in I$}
            \IF{$t = 1$}
                \STATE Sample $s_{t,i}$ from $\mu_i$
            \ELSE
                \STATE $s_{t,i} \gets P_t(s_{t-1,i}, a_{t-1,i})$
            \ENDIF
            \STATE $a_{t,i} \gets \pi_{i,t}(s_{t,i})$
        \ENDFOR
        \STATE Compute $\vec{r_t} \gets \vec{R}_t(\vec{s_t}, \vec{a_t})$
    \ENDFOR
    \STATE Compute $\vec{g} \gets \sum_{t=1}^T(\gamma^{t-1}\vec{r_t})$
    \STATE Compute $U \gets \prod_{i=1}^n(g_i + 1)$
    \STATE Define $C_{i,t}(s,a) := (\vec{R}_t(\vec{s_t}, \vec{a_t})$
    \STATE \hspace{1em} $- \vec{R}_t(\langle \vec{s_{t,-i}}, s \rangle, \langle \vec{a_{t,-i}}, a \rangle)) \cdot \frac{U \gamma^{t-1}}{\vec{g} + 1}$
    \STATE Sample agent $i \sim I$
    \STATE Optimize agent $i$ using $C$ and get $\pi^{*}_i$
    \STATE $\pi_i \gets \pi^{*}_i$
\ENDWHILE
\STATE \textbf{return} $\pi$
\end{algorithmic}
\end{algorithm}

See Appendix B for the pseudocode for the single-agent optimization subroutine. 

The time complexity of the entire algorithm is  $O(VNT|S||A|)$ where $V$ is the average number of optimizations per agent required for convergence. 

\subsection{Multi-agent Rollout Policy}

The Multi-agent Rollout algorithm, introduced by \citet{9317713}, extends the policy iteration (PI) framework to multi-agent systems. Rollout simplifies the optimization process by iteratively improving policies, optimizing each agent's actions sequentially while considering the fixed decisions of other agents. This sequential optimization significantly reduces computational complexity, making it particularly advantageous for large-scale multi-agent systems. By focusing on one agent at a time, the algorithm leverages the structure of the multi-agent MDP to approximate near-optimal solutions efficiently.

Traditional PI algorithms face scalability challenges in multi-agent environments due to the exponential growth in complexity when optimizing all agents' actions simultaneously. The Multi-agent Rollout method addresses this by performing local optimizations for each agent in a sequential manner, adjusting one agent's actions at a time while keeping the others fixed. As a result, the computational complexity scales linearly with the number of agents, \(n\), enhancing feasibility in systems with many agents.

This approach also supports real-time applications and dynamic replanning, where decisions must adapt to changing conditions continuously. By incorporating the rollout technique into multi-agent settings, we ensure that each agent optimizes its actions while maintaining coordination with others, ultimately producing globally improved policies that enhance system performance over successive iterations.

\subsubsection{Single-Agent Rollout Algorithm and Policy Improvement}

The standard one-step lookahead \textbf{single-agent} rollout algorithm is a method for improving a given base policy \( \pi = \{\pi_0, \dots, \pi_{T-1}\} \). The base policy refers to an initial policy that guides action selection before any optimization or improvement steps are applied. This base policy serves as a benchmark for evaluating and enhancing the rollout policy. At each time step \( t \), given the current state \( s_t \), \( Q_{t, \pi}(s_t, a_t) \) represents the expected cumulative reward of taking action \( a_t \) in state \( s_t \) and thereafter following the base policy \( \pi \):

The expected cumulative reward under the base policy \( \pi \), starting from state \( s_{t+1} \), is given by:

\begin{align}
    J_{t+1, \pi}(s_{t+1}) = \mathbb{E} \Bigg[ &\sum_{k = t+1}^{T-1} \gamma^{k-(t+1)} r_k(s_k, \pi_k(s_k)) + \notag \\
    &\gamma^{T-(t+1)} r_T(s_T) \Bigg]
\end{align}

\noindent In this equation:
\begin{itemize}
    \item The sum calculates the discounted rewards from time \( t+1 \) to the terminal time \( T \),
    \item \( r_T(s_T) \) is the terminal reward at time \( T \),
    \item The expectation is over the sequence of future states generated by following the policy \( \pi \).
\end{itemize}

The rollout policy \( \tilde{\pi} \) is then constructed by selecting, at each time step \( t \), the action \( \tilde{a}_t \) that maximizes the Q-factor. This is done by choosing:

\begin{equation}
    \tilde{a}_t = \operatorname{argmax}_{a_t \in A_t(s_t)} Q_{t, \pi}(s_t, a_t)
\end{equation}

By selecting actions that maximize \( Q_{t, \pi}(s_t, a_t) \), the rollout policy improves upon the base policy, resulting in a higher expected cumulative reward at each stage, as indicated by the inequality:

\begin{equation}
    J_{t, \tilde{\pi}}(s_t) \geq J_{t, \pi}(s_t) \quad \text{for all } s_t \text{ and } t.
\end{equation}

\subsubsection{Multi-Agent Rollout Algorithm with LPSolver Base Policy}
The \textbf{base policy} \( \pi \) can be derived using various methods that optimize decision-making for each agent. The methods should be designed to minimize the expected cumulative discounted cost and can be chosen based on the specific problem setting. In this work, we use \textsc{LPSolver} policy as the base policy. The policy leverages linear programming to solve each MDP efficiently. Further details on the specific implementation and mechanics of \textsc{LPSolver} policy are provided in Appendix \ref{app:Rollout}.

\paragraph{Calculation of \( J_{t+1, \pi} \) for Multiple MDPs} The expected future reward \( J_{t+1, \pi}(s_{t+1}) \) is directly obtained from the value function \( V_{i, \pi}(s_{t+1}) \), precomputed by \textsc{LPSolver}:
\begin{equation}
J_{t+1, \pi}(s_{t+1}) = V_{i, \pi}(s_{t+1}).
\end{equation}

\paragraph{Action Selection} The action \( \tilde{a}_{t,i} \) for agent \( i \) is selected by maximizing the expected cumulative reward over the horizon \( [t, T] \). For each action \( a_{t,i} \in A_{i}(s_t) \), a candidate action profile is constructed:

\begin{equation}
    \tilde{a}_{t,i} \in \operatorname{argmax}_{a_{t,i} \in A_{i}(s_t)} Q_{t, i}(s_t, \tilde{a}_{t,1}, \dots, a_{t,i}),
\end{equation}
where \( Q_{t, i}(s_t, \tilde{a}_{t,1}, \dots, a_{t,i}) \) is the expected cumulative reward starting from time \( t \).

The Q-factor for agent \( i \) is defined as:
\begin{equation}
    Q_{t, i}(s_t, \tilde{a}_{t,1}, \dots, a_{t,i}) = \mathbb{E} \left[ \sum_{k=t}^{T-1} \gamma^{k - t} r_k^i(s_k, a_k) \right],
\end{equation}

subject to:
\[
    a_k = 
    \begin{cases}
        (\tilde{a}_{t,1}, \dots, a_{t,i}, \pi_k^{i+1}(s_k), \dots, \pi_k^n(s_k)), & \text{if } k = t, \\[8pt]
        (\pi_k^1(s_k), \dots, \pi_k^n(s_k)), & \text{if } k > t.
    \end{cases}
\]

\paragraph{Time Complexity} The computational complexity of this approach is \( O(T^2 N^2 |A| b) \), where \( T \) is the planning horizon, \( N \) is the number of agents, \( |A| \) is the action space size, and \( b \) is the complexity of calling the base policy's \textsc{get\_next\_action()} function. This reflects the iterative, multi-agent optimization process, balancing higher computational demands with robust policy improvement.

\begin{algorithm}[H]
\caption{Multi-Agent Rollout Algorithm with LPSolver Base Policy}
\label{alg:MAR_LP}
\textbf{Input}: Initial state \( s_0 \), Base policy \( \pi = \{\pi_0, \dots, \pi_{T-1}\} \) (implemented via \textsc{LPSolver} with discount factor \( \gamma \)), Horizon \( T \), State transition function \( P_t(s_t, a_t) \), Reward function \( r_t(s_t, a_t) \)\\
\textbf{Output}: \textsc{Rollout} policy \( \tilde{\pi} = \{\tilde{\pi}_0, \dots, \tilde{\pi}_{T-1}\} \)

\begin{algorithmic}[1]
\FOR{$t = 0$ to $T-1$}
    \STATE Observe current state \( s_t \)
    \FOR{each agent $i = 1$ to $n$}
        \STATE Initialize \( V_{\text{best}} = -\infty \)
        \FOR{each action $a_{t, i} \in A_{i}$}
            \STATE Set actions for agents $1 \leq j \leq i-1$:
                $a_{t, j} \gets \tilde{a}_{t,j}$
            \STATE Set current action for agent \( i \): $a_{t,j} =\gets a$
            \STATE Set actions for agents $i+1 \leq j < n$: 
                $a_{t,j} \gets \pi_t(s_{t,i})$
            \STATE Compute next state:
            \[
            s_{t+1} = P_t \left( s_t, \tilde{a}_{t,1}, \dots, a_{t,i}, \pi_{t,i+1}(s_t), \dots, \pi_{t,n}(s_t) \right)
            \]
            \STATE Simulate base policy \( \pi \) from \( s_{t+1} \) to \( T \) to get \( J_{t+1, \pi}(s_{t+1}) \)
            \STATE Calculate Q-value:
            \[
            Q = r_t \left( s_t, \tilde{a}_{t,1}, \dots, a_{t,i}, \pi_{t,i+1}(s_t), \dots, \pi_{t,n}(s_t) \right) + 
            \]
            \[
            \gamma J_{t+1, \pi}(s_{t+1})
            \]
            \IF{$Q > V_{\text{best}}$}
                \STATE Update \( V_{\text{best}} = Q \) and \( \tilde{a}_{t,i} = a_{t,i} \)
            \ENDIF
        \ENDFOR
        \STATE Update \textsc{Rollout} policy \( \tilde{\pi}_{t,i}(s_t) = \tilde{a}_{t,i} \)
    \ENDFOR
    \STATE Execute joint action \( \tilde{a}_t = (\tilde{a}_{t,1}, \dots, \tilde{a}_{t,n}) \)
    \STATE Transition to next state \( s_{t+1} = P_t(s_t, \tilde{a}_t) \)
\ENDFOR
\STATE \textbf{return} \( \tilde{\pi} = \{\tilde{\pi}_0, \dots, \tilde{\pi}_{T-1}\} \)
\end{algorithmic}
\end{algorithm}

\section{Experiments}

In this section, we run simulations to address the challenge of optimizing crop planning for small-scale farmers in India, where traditional systems often lead to income loss by failing to account for real-time market supply and demand. In this study, we evaluate three different reinforcement learning policies (\textsc{IQL}, \textsc{ABA}, and \textsc{Rollout}) to measure how effectively they perform in a multi-agent crop planning environment. We test their impact on runtime, reward distribution, discount factors, and slope coefficients. All policies are aimed at maximizing total farmer income while allowing us to observe fairness outcomes among agents. Testing was conducted on a Ryzen 5600H 3.0 GHz processor, without GPU, and the code was written in Python 3.9.

Our first experiment was total joint reward, which is the sum of farmer rewards. Experiments were set up with 5, 10, 15, and 20 agents, and we measured how much total reward each policy generated. Unless otherwise noted, 14 days passes with each timestep for a total of $T=26$ timesteps. We set the $\Delta_R$ parameter to 0.1 and the slope coefficient to 500.

\begin{figure}[ht]
    \centering
    \includegraphics[width=8cm]{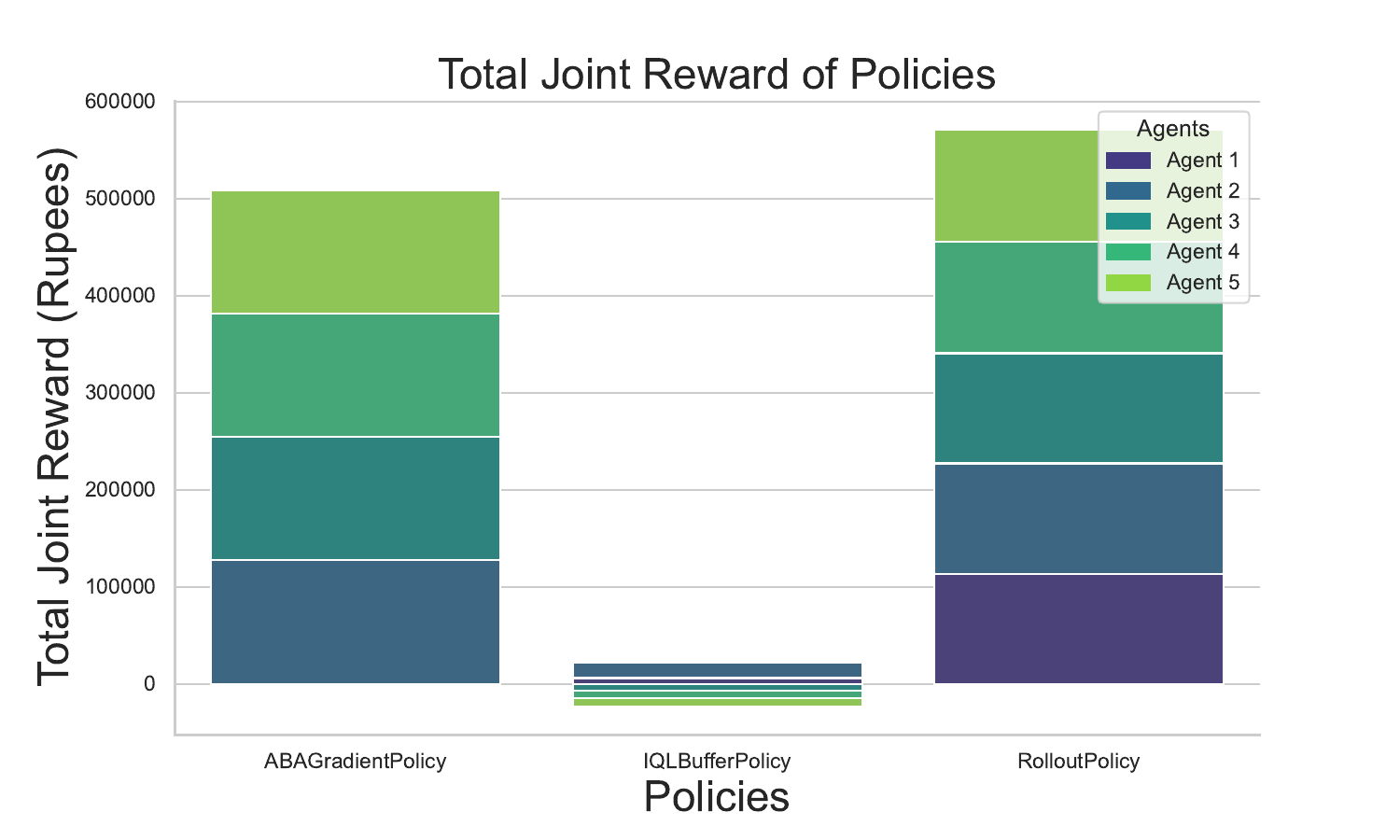}
    \caption{Total Joint Reward of Policies}
    \label{fig:jointreward}
\end{figure}

The results showed that both \textsc{ABA} and \textsc{Rollout} produced significantly higher total rewards, each reaching approximately 500,000 rupees, while \textsc{IQL} achieved much lower rewards. For \textsc{Rollout}, the rewards per agent were relatively consistent, with each agent making around 100,000 rupees, which indicates a strong distributional balance. \textsc{ABA} also showed relatively fair reward distribution but had one agent with zero rewards, suggesting only partial fairness. Here, we define ``fairness" as equal distribution of rewards across agents, a common metric for equity in MARL applications. This concept of fairness has been explored by researchers as a benchmark to measure equitable outcomes among agents, particularly in settings where agents are interdependent \citep{ju2024achieving}.

\textsc{IQL} was the least effective in reward distribution, with some agents even experiencing negative rewards. This result reflects IQL’s non-stationarity, where each agent optimizes based solely on its individual states and rewards, while other agents' actions continuously alter the environment. As a result, agents lack sufficient time to learn and converge on optimal joint actions. In contrast, \textsc{ABA} uses dynamic programming to work backward from the horizon, capturing the true value of each state with a comprehensive cost function. \textsc{Rollout}, by using an LPSolver as its base, shows some limitations; since the LPSolver optimizes for single-agent policies, \textsc{Rollout} mostly coordinates all agents to take similar actions. While this alignment maximizes reward within the simulation’s parameters, it risks oversaturating the market if applied without such constraints in practice.

\begin{figure}[ht]
    \centering
        \includegraphics[width=8cm]{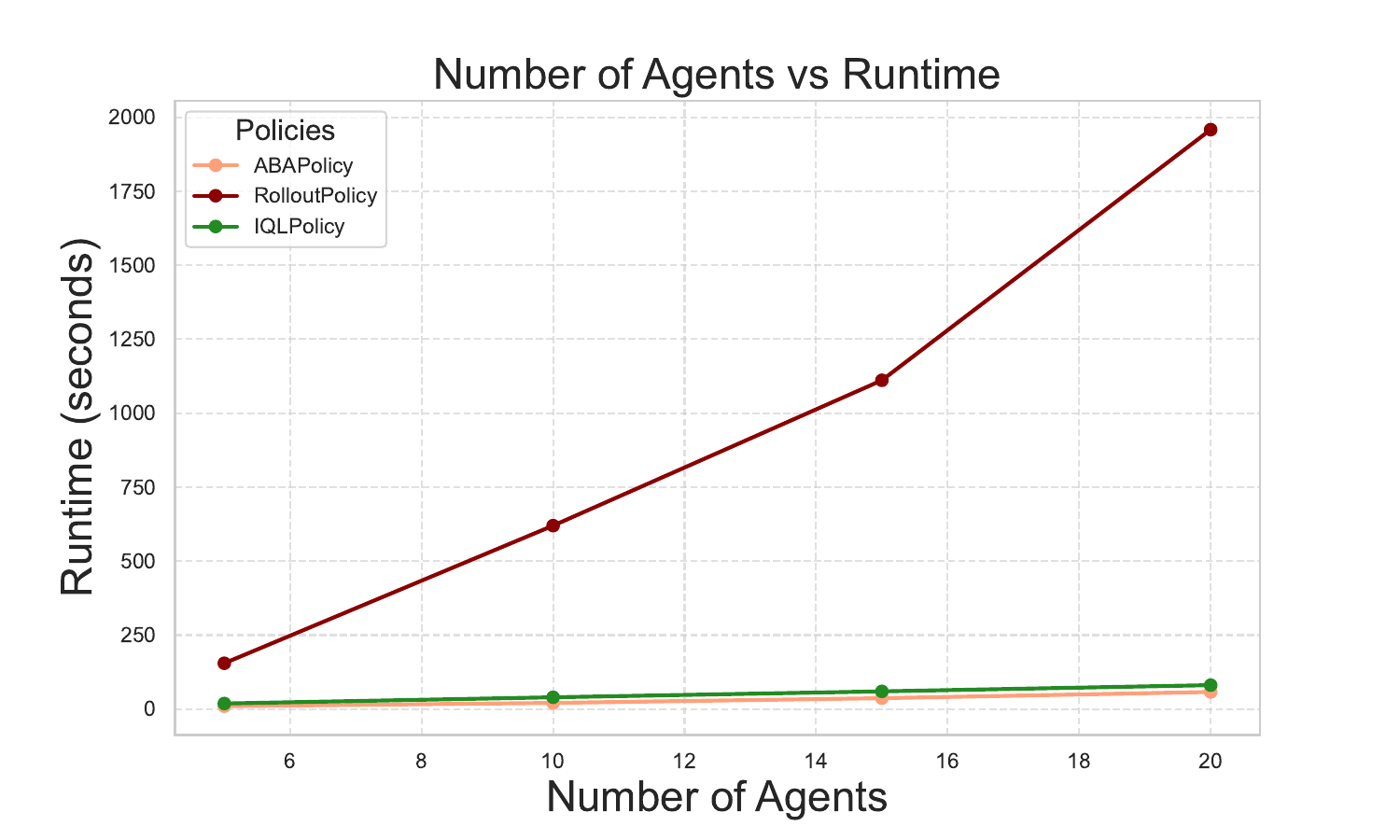}
    \caption{Number of Agents vs. Runtime for Different Policies}
    \label{fig:runtime}
\end{figure}

For the runtime analysis, we used a general MDP model to avoid the complexities of greenhouse-specific details. This allowed us to focus on core performance and scalability, providing a clearer and faster comparison of how each policy handles multiple agents without the added complexity of the greenhouse constraints, such as crop maturity, market interactions, or seasonal factors.

The simulation results indicate that \textsc{Rollout} required significantly more runtime as the number of agents increased, reaching over 3500 seconds for 20 agents, as shown in Figure \ref{fig:runtime}. This extended runtime makes \textsc{Rollout} less practical for large-scale applications. In contrast, \textsc{IQL} and \textsc{ABA} completed their tasks much faster. \textsc{IQL} exhibited \( O(N) \) time complexity, scaling linearly with the number of agents. \textsc{ABA} also showed a manageable runtime with \( O(N \times S) \) time complexity, where \( S \) is the number of states, reflecting its stepwise optimization of each agent’s policy through dynamic programming. This complexity is consistent with the results shown, where ABA scales efficiently with increasing agent numbers while balancing computational demand. In summary, while \textsc{Rollout} achieves the highest rewards, its larger time complexity limits scalability. \textsc{IQL} is efficient but sacrifices reward performance, whereas \textsc{ABA} achieves a balance of decent rewards with a reasonable runtime.

\begin{figure}[ht]
    \centering
    \includegraphics[width=8cm]{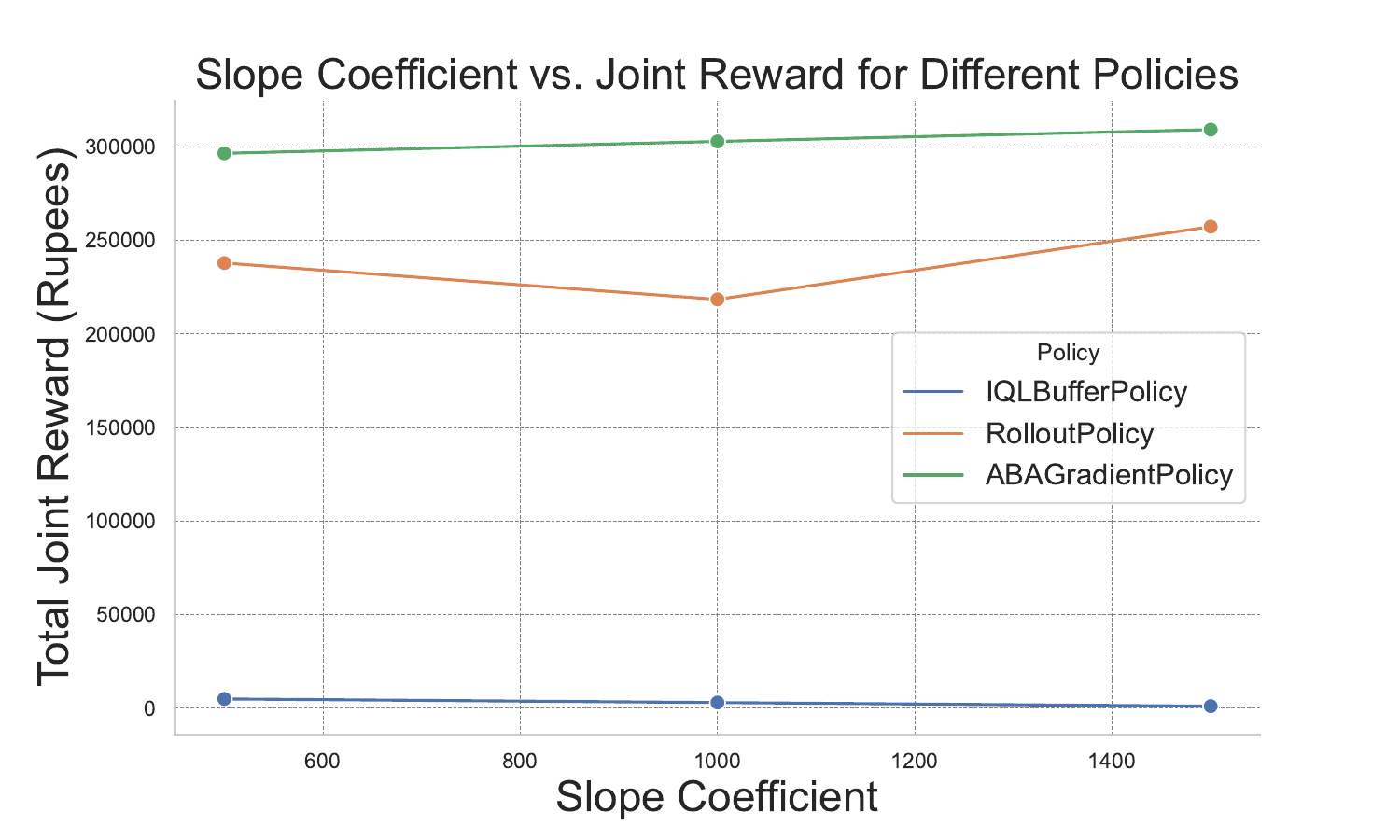}
    \caption{Total Joint Reward vs. Slope coefficients for Different Policies}
    \label{fig:slope}
\end{figure}

Finally, we examined the effect of different slope coefficients—values ranging from 500 to 1,500—on the total joint reward across 26 horizons, 14 timesteps, and 2 agents. The slope coefficient reflects how sensitive the market is to oversupply; as more agents harvest the same crop, prices drop more sharply with higher slope values. This coefficient directly impacts total reward: if agent \(i\) follows the same policy but under different slope coefficients, the total reward can vary because the market price adjusts differently depending on supply sensitivity. Therefore, we interpret each point on the plot in light of this supply sensitivity.

With \textsc{IQL}, the reward remains flat across all slope coefficients, showing that IQL’s independent agent actions make it unable to adapt to supply-driven price sensitivity. This flat result indicates that the slope coefficient does not influence IQL’s total reward outcome. By contrast, \textsc{ABA} shows a steady increase in reward as the slope coefficient rises. This upward trend suggests that \textsc{ABA} can effectively adapt to market conditions, likely by spreading out agents’ harvest times to avoid market oversaturation. \textsc{Rollout}, meanwhile, displays a slight dip around a slope coefficient of 1,000, possibly due to challenges in coordinating actions at this particular sensitivity. However, \textsc{Rollout} recovers as the slope coefficient increases, which indicates that it can adjust agents’ actions to balance market supply and demand at higher sensitivity values. Overall, \textsc{ABA} and \textsc{Rollout} better handle market dynamics by leveraging coordination, making them more effective in scenarios where agents must balance their actions to avoid oversupply and maintain favorable prices.
 
\begin{figure}[ht]
    \centering
    \includegraphics[width=8cm]{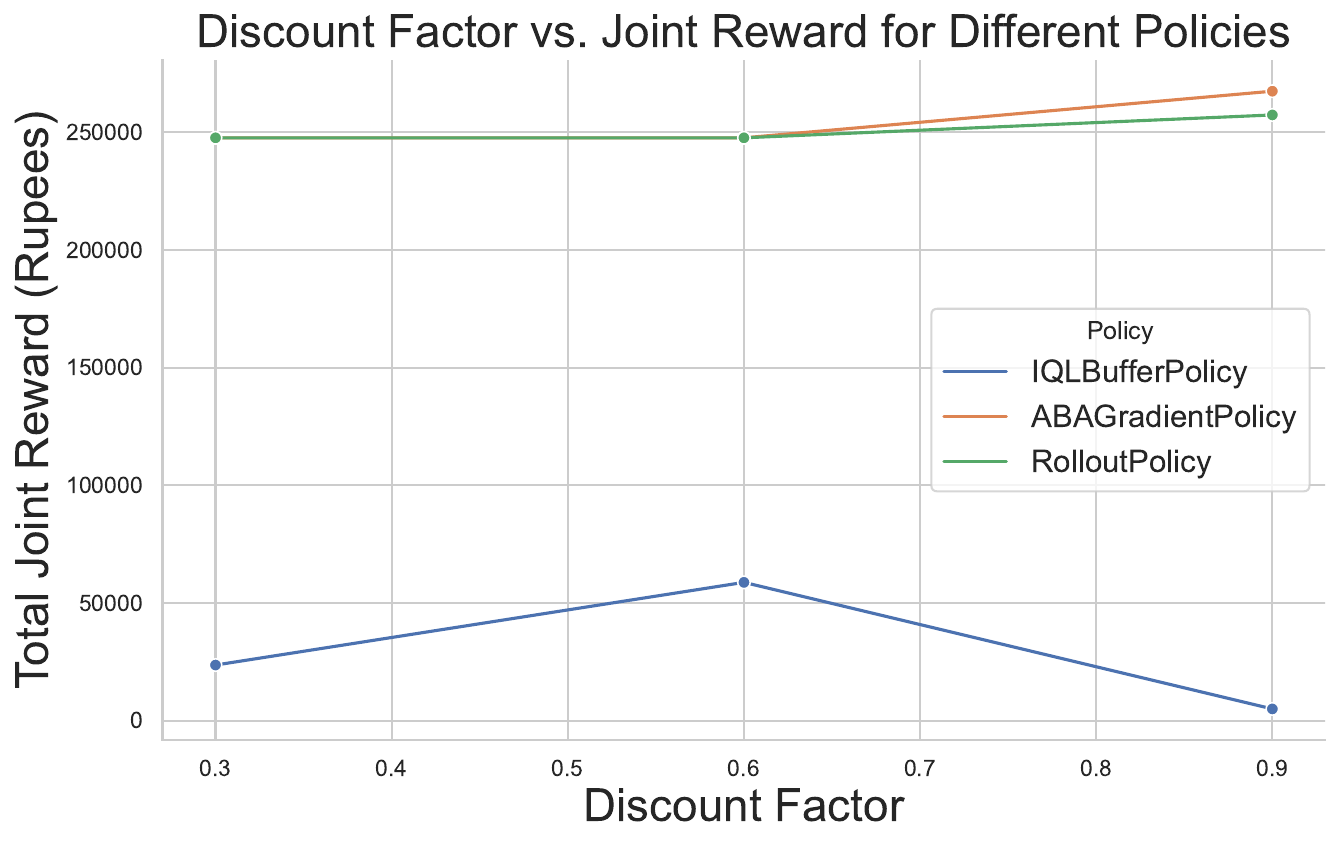}
    \caption{Total Joint Reward vs. Discount Factors for Different Policies}
    \label{fig:discount}
\end{figure}

Next, we observed how discount factors (ranging from 0.3 to 0.9) impact the total joint reward for the three policies across 26 horizons, 14 timesteps, and 2 agents. A discount factor represents how much weight the policies place on future rewards compared to immediate ones. \textsc{ABA} and \textsc{Rollout} maintain consistently high rewards, with a slight upward trend as the discount factor increases. These policies, which optimize for joint rewards, are better at managing the long-term impact of actions and coordinating between agents, ensuring that they continue to maximize total rewards even as future rewards become more important. In contrast, \textsc{IQL} reaches its highest performance at a discount factor of around 0.6 but then drops off as the discount factor increases. This behavior likely occurs because IQL operates independently for each agent, meaning that agents do not coordinate their actions. As the discount factor increases, agents focus more on long-term rewards, but since they are not considering the actions of other agents, their decisions negatively affect overall outcomes. This leads to lower joint rewards, particularly at higher discount factors.

\section{Conclusion}

In this study, we explored the distinct trade-offs among computational efficiency, reward optimization, and fairness in multi-agent crop planning using different MARL algorithms. IQL initially appeared advantageous due to its computational efficiency; however, as the number of agents increases, IQL's performance declines due to coordination challenges, resulting in lower overall rewards and a less equitable outcome distribution. This difficulty with coordination limits IQL's suitability for multi-agent contexts where fairness is a priority.

The ABA policy, which plans actions for each agent in advance, offers a compromise between coordination and computational demand. Unlike IQL, ABA enhances agent coordination by optimizing sequentially, which improves fairness and reward distribution. However, its moderate computational overhead must be weighed against its performance benefits.

Conversely, the multi-agent \textsc{Rollout} policy achieves the highest rewards and equitable outcomes by optimizing joint agent actions simultaneously. This approach, however, significantly increases runtime, making it less feasible for large-scale applications. This trade-off implies that \textsc{Rollout} may only be practical in scenarios with fewer agents or lower computational constraints.

The experimental analysis was crucial in clarifying these trade-offs, providing insights into each algorithm's strengths and limitations under varying conditions. Ultimately, these findings underscore that selecting an appropriate MARL approach for agricultural decision-support systems requires balancing efficiency, reward optimization, and fairness according to specific needs. To address the scalability of these algorithms, we recommend further exploration into enhancing MARL policies that can balance fairness and efficiency in resource-constrained environments, ensuring sustainable support for farmers.

\bibliography{main}

\newpage
\appendix
\onecolumn
\section{Details On Linear Approximation for Agent-by-Agent Algorithm}
\label{app:AbA_linear}

Here, we will cover in more detail how we obtain a linear approximation for the effect of a change in a single reward for a single agent. Specifically, we approximate the change in $U$ under some change $r_{t,j} \leftarrow r_{t,j} + \epsilon$ as $\epsilon * \frac{\partial U}{\partial r_{t,j}} |_{\pi}$. 

We expand out the partial derivative using the multi-variable chain rule. Since $r_{t,j}$ only has an effect on $g_j$, we are left with only one term. We can calculate each of these partial derivatives by consulting their definition equations, and can then combine the entire linear approximation into one concise equation. Again, we will use $\vec{r}_t, \vec{g},$ and $U$ values derived from previous values of $\pi$.

\begin{align}
    \frac{\partial U}{\partial r_{t,j}} &= \frac{\partial U}{\partial g_j} \cdot \frac{\partial g_j}{\partial r_{t,j}} \\
    \frac{\partial U}{\partial g_j} &= \prod_{1 \leq i \leq n, i \neq j}(g_i+1) = \frac{U}{g_j+1} \\
    \frac{\partial g_j}{\partial r_{t,j}} &= \gamma^{t-1} \\
    \frac{\partial U}{\partial r_{t,j}} &= \frac{U}{g_j+1} \cdot \gamma^{t-1}
\end{align}

Now, we have an approximation for the change in $U$ based on some change in $r_{t,j}$ for some agent $j$ and timestep $t$. We now must turn this into an approximation for the instantaneous change in $U$ based on some change in $a_{t,i}$ and $s_{t,i}$ for some $t$. We do this by calculating the effect of the action change on $r_{t,}$ for all agents $j$, and multiplying each of these terms by the previously calculated effect of $r_{j,t}$ on $U$, and then taking the sum of these products. We can then manipulate the algebra, to derive $C_{i,t}(s, a)$ defined as the the effect on $U$ of changing $s_{i,t} \leftarrow s$ and $a_{i,t} \leftarrow a$.

\begin{align}
C_{i,t}(s,a) &=  \sum_{j=1}^{n} \left( \vec{R}_t(\vec{s_t}, \vec{a_t})_j - \vec{R}_t(\langle \vec{s_{t,-i}}, s \rangle, \langle \vec{a_{t,-i}}, a \rangle)_j \right) \frac{\partial U}{\partial r_{t,j}} \\
&= \left( \vec{R}_t(\vec{s_t}, \vec{a_t}) - \vec{R}_t(\langle \vec{s_{t,-i}}, s \rangle, \langle \vec{a_{t,-i}}, a \rangle) \right) \cdot 
   \begin{pmatrix} 
      \frac{\partial U}{\partial r_{t,1}} \\[6pt] 
      \frac{\partial U}{\partial r_{t,2}} \\[6pt] 
      \vdots \\[6pt] 
      \frac{\partial U}{\partial r_{t,n}} 
   \end{pmatrix} \\ 
&= \left( \vec{R}_t(\vec{s_t}, \vec{a_t}) - \vec{R}_t(\langle \vec{s_{t,-i}}, s \rangle, \langle \vec{a_{t,-i}}, a \rangle) \right) \cdot 
   \begin{pmatrix} 
      \frac{U}{g_1 + 1} \gamma^{t-1} \\[6pt] 
      \frac{U}{g_2 + 1} \gamma^{t-1} \\[6pt] 
      \vdots \\[6pt] 
      \frac{U}{g_n + 1} \gamma^{t-1} 
   \end{pmatrix} \\
&= \left( \vec{R}_t(\vec{s_t}, \vec{a_t}) - \vec{R}_t(\langle \vec{s_{t,-i}}, s \rangle, \langle \vec{a_{t,-i}}, a \rangle) \right) \cdot \frac{U \gamma^{t-1}}{\vec{g} + 1}
\end{align}

The above expression is now dependent only on the state and action trajectories, which can be simulated using the previous version of the joint-policy before computing $DP$ values. Putting this altogether, we get the $DP$ transitions we need.

\begin{equation}
    DP_t(s) = \max_a \left( C_{i,t}(s,a) + DP_{t+1}(P_t(s,a)) \right)
\end{equation}

As a base case, $DP$ values for $t = T$ should have no term represent future welfare improvement, because the MDP is at the last step of the horizon. The transitions here are thus simply:

\begin{equation}
    DP_T(s) = \max_a \left( C_{i,T}(s,a) \right)
\end{equation}

Once we have calculated all $DP$ values, we can easily recover the single-agent-optimized policy by from each timestep choosing the action that yielded the best improvement according to the transition equation. Specifically, we will take

\begin{equation}
    \pi_{i,t}(s) = \underset{a}{\operatorname{argmax}} \left( C_{i,t}(s,a) + DP_{t+1}(P_t(s,a)) \right)
\end{equation}

\section{Pseudocode For Single-Agent Optimization Subroutine in Agent-by-Agent Algorithm}

The following pseudocode is the implementation of line 18 in Algorithm 2.

\begin{algorithm}[] 
\caption{Single-Agent Optimization Method}
\label{alg:AAO_DP1}
\textbf{Input}: agent $i$, linear approximator $C$ \\
\textbf{Output}: Optimized policy for agent $i$ $\pi_i$

\begin{algorithmic}[1]
\STATE Initialize $DP_t(s) \gets 0$ for all $t \in \{1, \dots, T\}$, $s \in S$
\FOR{$t = T$ to $1$}
    \FOR{$s \in S$}
        \IF{$t = T$}
            \STATE $DP_t(s) \gets \max_a (C_{i,t}(s,a))$
            \STATE $\pi_{i,t}(s) \gets \operatorname{argmax}_a (C_{i,t}(s,a))$
        \ELSE
            \STATE $DP_t(s) \gets \max_a (C_{i,t}(s,a) + DP_{t+1}(P_t(s,a)))$
            \STATE $\pi^{*}_{i,t}(s) \gets \operatorname{argmax}_a (C_{i,t}(s,a) + DP_{t+1}(P_t(s,a)))$
        \ENDIF
    \ENDFOR
\ENDFOR
\STATE \textbf{return} $\pi^{*}_i$
\end{algorithmic}
\end{algorithm}

\section{Details of LPSolver Base Policy}
\label{app:Rollout}

\paragraph{LPSolver Base Policy for Multiple MDPs.} The \textsc{LPSolver} policy is a specific implementation of the base policy, solving each MDP as a linear program to minimize expected cumulative discounted cost. The value function \( V^\pi_i(s) \) for each state \( s \) and agent \( i \) is obtained by solving:

\begin{align}
\text{Minimize} \quad & \sum_{s \in S_i} V_i(s) \\
\text{Subject to} \quad & V_i(s) \geq r_i(s, a) + \gamma \sum_{s' \in S_i} P_i(s' \mid s, a) V_i(s'), \quad \forall s, a.
\end{align}

The stationary policy \( \pi_i(s) \) is then:

\begin{equation}
\pi_i(s) = \arg\max_{a \in A_i(s)} \left[ r_i(s, a) + \gamma \sum_{s'} P_i(s' \mid s, a) V_i(s') \right].
\end{equation}

\end{document}